\documentclass[sigconf]{acmart}
\AtBeginDocument{%
  }

\copyrightyear{2026}
\acmYear{2026}
\setcopyright{cc}
\setcctype{by}
\acmConference[WWW '26]{Proceedings of the ACM Web Conference 2026}{April 13--17, 2026}{Dubai, United Arab Emirates}
\acmBooktitle{Proceedings of the ACM Web Conference 2026 (WWW '26), April 13--17, 2026, Dubai, United Arab Emirates}
\acmPrice{}
\acmDOI{10.1145/3774904.3793028}
\acmISBN{979-8-4007-2307-0/2026/04}

\usepackage{amssymb}

\usepackage[table,x11names]{xcolor}
\usepackage{color, colortbl}
\definecolor{LightCyan}{rgb}{0.88,1,1}
\definecolor{maroon}{cmyk}{0,0.87,0.68,0.32}
\usepackage{tabularray}
\usepackage{balance}

\begin{document}

\title{Mining Citywide Dengue Spread Patterns in Singapore Through Hotspot Dynamics from Open Web Data}


\author{Liping Huang}
\affiliation{%
  \institution{Agency for Science, Technology and Research (A*STAR)}
  \country{Singapore}
}
\email{liping.huang.sg@gmail.com}

\author{Gaoxi Xiao}
\authornote{Corresponding authors}
\affiliation{%
  \institution{Nanyang Technological University}
  \country{Singapore}
  }
\email{egxxiao@ntu.edu.sg}

\author{Stefan Ma}
\affiliation{%
  \institution{Ministry of Health, Singapore}
  \country{Singapore}}
\email{stefan_ma@moh.gov.sg}

\author{Hechang Chen}
\authornotemark[1]
\affiliation{%
  \institution{Jilin University}
  \city{Changchun}
  \country{China}
}
\email{chenhc@jlu.edu.cn}

\author{Shisong Tang}
\affiliation{
    \institution{Tsinghua University}
    \city{Beijing}
    \country{China}
}
\email{tangshisong13@gmail.com}

\author{Flora Salim}
\affiliation{%
  \institution{University of New South Wales}
  \city{Sydney}
  \state{NSW}
  \country{Australia}
  }
\email{flora.salim@unsw.edu.au}

\renewcommand{\shortauthors}{Liping Huang et al.}

\begin{abstract}
Dengue, a mosquito-borne disease, continues to pose a persistent public health challenge in urban areas, particularly in tropical regions such as Singapore. Effective and affordable control requires anticipating where transmission risks are likely to emerge so that interventions can be deployed proactively rather than reactively. 
This study introduces a novel framework that uncovers and exploits \textbf{latent transmission links} between urban regions, mined directly from publicly available dengue case data. Instead of treating cases as isolated reports, we model how hotspot formation in one area is influenced by epidemic dynamics in neighboring regions. While mosquito movement is highly localized, \textbf{long-distance transmission is often driven by human mobility}, and in our case study, the learned network aligns closely with commuting flows, providing an interpretable explanation for citywide spread. These hidden links are optimized through gradient descent and used not only to \textbf{forecast hotspot status} but also to \textbf{verify the consistency of spreading patterns}, by examining the stability of the inferred network across consecutive weeks.
Case studies on Singapore during 2013–2018 and 2020 show that four weeks of hotspot history are sufficient to achieve an average F-score of 0.79. Even under the COVID-19 “circuit breaker,” when mobility patterns were severely disrupted, the model remained robust with an F-score of 0.83. Importantly, the learned transmission links align with commuting flows, highlighting the interpretable interplay between hidden epidemic spread and human mobility.
By shifting from simply reporting dengue cases to \textbf{mining and validating hidden spreading dynamics}, this work transforms open web-based case data into a predictive and explanatory resource. The proposed framework advances epidemic modeling while providing a scalable, low-cost tool for public health planning, early intervention, and urban resilience.

\end{abstract}

\ccsdesc[500]{Applied computing~Life and medical sciences}
\ccsdesc[300]{Computing methodologies~Modeling and simulation}
\ccsdesc[300]{Computing methodologies~Data mining}
\ccsdesc[100]{Computing methodologies~Benchmarking}

\keywords{Dengue Cases, Disease Spreading Pattern, Hotpot Dynamics, Machine Learning}


\maketitle

\section{Introduction}
Dengue, a vector-borne viral infection transmitted to human kinds mainly by Aedes aegypti \cite{WHO2024, Mariana2023}, remains endemic in many tropical and subtropical regions \cite{b2}. Each year, an estimated 60–100 million symptomatic infections and 14,000–20,000 deaths are attributed to dengue \cite{b3,b4}, with an annual global health cost estimated at about 9 billion dollars \cite{b5}. Dengue virus is grouped into four serotypes (DENV 1-4), which are closely related to each other, yet antigenically distinct and genetically diverse \cite{b6}. Limited understanding of the immunological interactions among these serotypes—particularly those that exacerbate disease severity—has hindered the development of effective and widely available vaccines \cite{b7}. Consequently, vector control through insecticides and larval source reduction remains the primary strategy for dengue prevention \cite{b8}.

Singapore, an equatorial city-state, faces heightened vulnerability due to its humid climate and highly urbanized environment, both of which provide ideal conditions for Aedes breeding and dengue transmission \cite{b10}. All four dengue serotypes co-circulate year-round \cite{b11}, contributing to recurring outbreaks and occasional nationwide epidemics \cite{b12,b13}. Major epidemics were recorded in 2013, 2014, and 2020, with the 2013 outbreak reaching 404.9 cases per 100,000 population (22,170 total cases) \cite{b15}, and the unprecedented 2020 outbreak reporting 35,315 cumulative cases \cite{b14}. Resurgence was also noted in 2022, with 1,104 cases in a single week ending 30 July \cite{b16}.

Dengue imposes a significant economic burden in Singapore, estimated at US\$42.5 million annually \cite{b17}. To mitigate risk, the National Environment Agency (NEA) conducts regular inspections, eliminates breeding sites, and engages the public in preventive measures \cite{b18,b19}. During the 2020 outbreak alone, NEA carried out approximately 107,000 home inspections in May and June \cite{b14}. While effective, these large-scale efforts are resource-intensive. Their efficiency could be greatly improved by focusing interventions on areas at higher risk of future outbreaks. This requires methods that can anticipate where dengue hotspots will emerge \cite{b20}.

Existing research has explored dengue forecasting from multiple angles. Climatic variables such as rainfall and temperature have been used to predict national-level incidence weeks in advance \cite{b23}, but these approaches lack fine-grained spatial resolution. Other studies integrate diverse urban datasets—including population density, vector surveillance, vegetation indices, meteorology, and mobility data—together with machine learning models such as random forests to stratify spatial risks \cite{b24,b25}. While informative, these methods often require extensive, fine-grained, and near–real-time data (e.g., mobile phone mobility or detailed epidemiological records), which may not be feasible in many dengue-endemic settings \cite{b8}. 
The study in \cite{b8} proposes to pick up 32 grids (1km × 1km) with highest numbers of predicted dengue cases. Such a forecasting certainly helps anchor urban regions that are with the highest dengue transmission risks. The method however demands rich datasets as inputs, including the residential address and date of onset of each confirmed case, the movement patterns recognized from mobile phone data, age of buildings, and fine-grained meteorological data etc., all of which need to be available almost in real time. Access to such rich data in real time however may not be feasible in some dengue endemic cities or nations, which impedes the applicability of the method. Despite of the extensive research efforts, a practical method for dynamically forecasting dengue risk at the urban scale, relying only on epidemiological surveillance data, remains a critical gap.

In Singapore, epidemiological data are systematically published by the National Environment Agency on a weekly basis and made publicly accessible through the \hyperlink{https://outbreak.sgcharts.com/data}{
SGCharts platform}. Leveraging this open dataset, our objective is to develop a method that uncovers spatial spreading patterns of dengue hotspots from epidemiological inputs alone. While mosquito movement is highly localized, long-distance transmission can arise through human mobility. In our case study, we show that the learned spreading network not only captures these unobserved dynamics but also exhibits strong alignment with the human mobility network, providing an interpretable explanation of how dengue can spread across the city. This approach offers government agencies a practical tool to enhance vector control planning, proactively allocate resources, and ultimately mitigate dengue transmission.

\section{Graphical Model View of Dengue Hotspot} 

\begin{figure}[ht]
    \centering
    \includegraphics[width=0.15\textwidth]{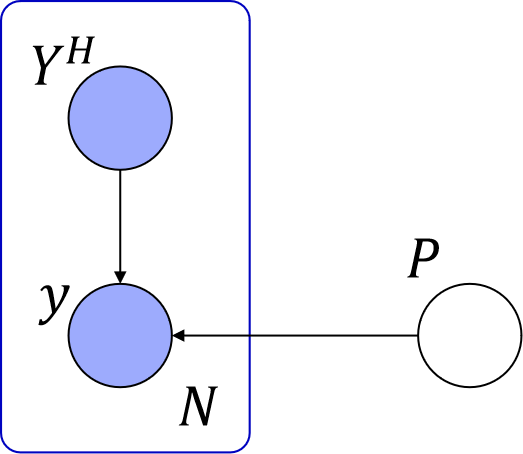}
    \caption{DAG for the Dengue Hotspot Spreading.}
    \label{fig:dag}
\end{figure}

We represent the dynamics of dengue hotspot spreading across urban regions using a directed acyclic graph (DAG) as in Fig. \ref{fig:dag}. The current-week hotspot $y$ in a region is influenced by two factors: (i) the observed hotspots of all regions in the previous $H$ weeks, denoted $\mathbf{Y}^{H}$, and (ii) the hidden pattern of transmission links across urban regions, $P$. 

The figure illustrates the two dominant causal pathways considered in our analysis: temporal persistence of hotspots ($\mathbf{Y}^{H} \to y$) and human-mediated long-distance transmission links ($P \to y$). 
We intentionally omit explicit ecological factors that determine mosquito abundance. This choice is based on two considerations. First, mosquito activity and local ecological conditions primarily affect transmission within short spatial neighborhoods. Second, the previous $H$-week hotspot matrix $\mathbf{Y}^{H}$ effectively summarizes these local effects, as recent infections already reflect the influence of the ecological factors. Including them as separate variables would add complexity without providing additional causal information for recognizing cross-region spread.

By focusing on $\mathbf{Y}^{H}$ and $P$, the model captures the key mechanisms for citywide dengue transmission while remaining parsimonious and interpretable. This structure is well-suited for mining latent inter-region transmission links and emphasizes the pathways most relevant at the urban scale, while implicitly accounting for local entomological and ecological influences.

This simplification also improves interpretability and identifiability. Including ecological or entomological factors risks conflating local dynamics with cross-region transmission, while in practice these variables are often noisy or unavailable at the required resolution. By retaining only the two strongest causal channels—autoregressive persistence ($Y^{H} \to y$) and unobservale spatial transmission link ($P \to y$)—we focus the model squarely on uncovering the latent transmission network across the city, which is the central objective of this study.

\begin{figure*}[t]
    \centering
    \includegraphics[width=0.9\textwidth]{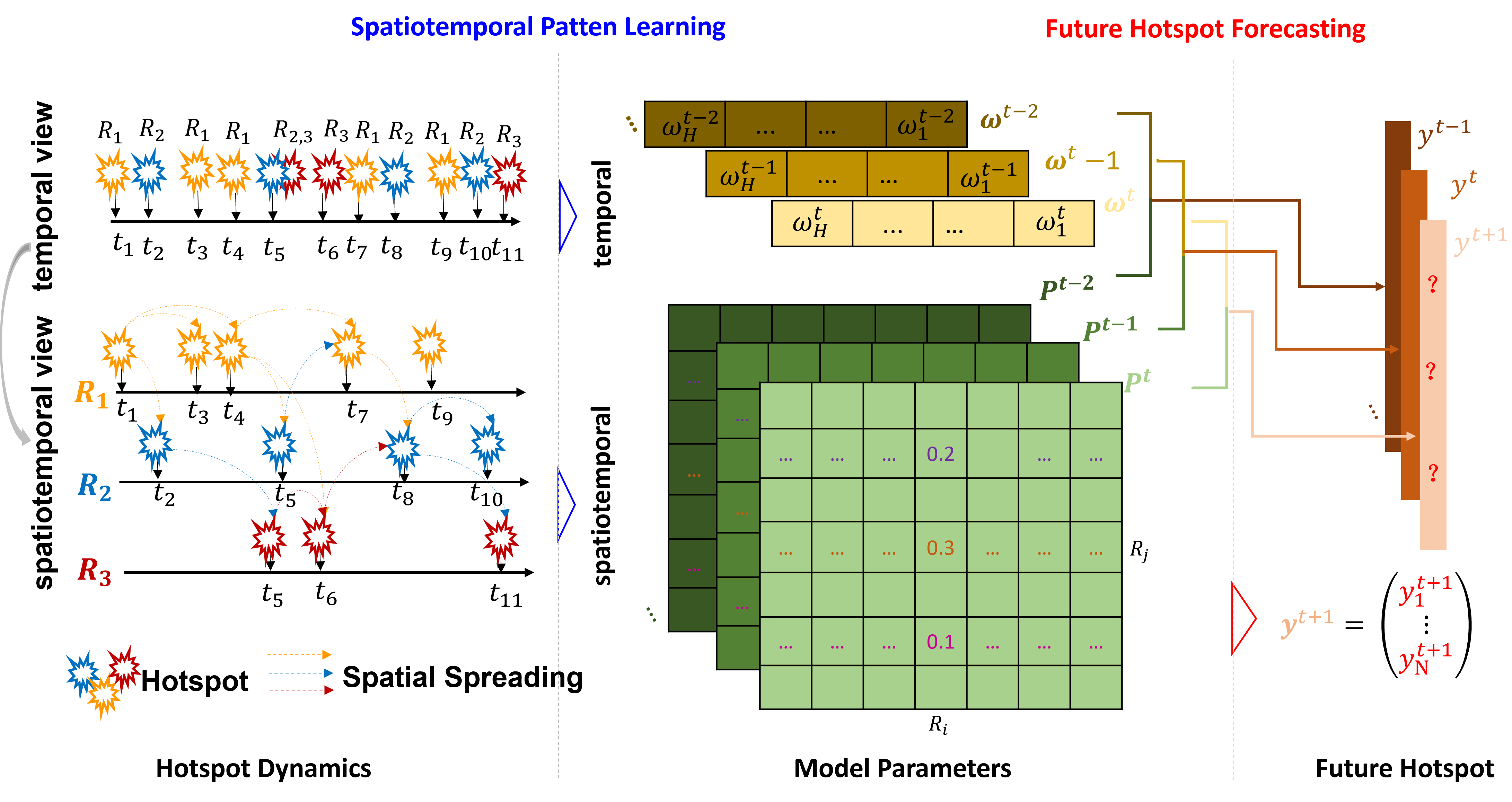}
    \caption{Diagram of Dengue Hotspot Dynamics Modeling with Spatiotemporal Patten Learning and Hotspot Forecasting.}
    \label{fig:overview}
\end{figure*}

\section{Methodology}
\subsection{Dengue Case Data}
The NEA of Singapore regularly publishes geographical dengue transmission localities, each of which is the residential apartment or the workplace address of an infected person. This study uses dengue case data collected and published by SGCharts Outbreak (\url{https://outbreak.sgcharts.com/data}
), which systematically archived publicly available information from the National Environment Agency Singapore. We focus on the machine-friendly CSV files from 2013 to 2018 and 2020, capturing weekly dengue dynamics at high spatial granularity.

Each CSV file corresponds to a snapshot of NEA’s dengue cluster webpage and includes structured fields for street address, latitude, longitude, cluster number, number of recent cases (last 2 weeks), total cluster cases, and collection date. Geocoded locations allow precise mapping of dengue cases at the apartment block level, enabling spatial and temporal analysis of outbreak patterns. Filenames follow a YYMMDD convention to indicate the date of data capture, ensuring chronological alignment for longitudinal studies. Historical coverage is particularly valuable because NEA only publishes current dengue data on its website; SGCharts preserves these historical snapshots, facilitating retrospective analyses and predictive modeling.

\subsection{Dengue Hotspot}
 In this study, the CSV datasets in the years from 2013 to 2018, and in the unprecedent dengue outbreak year 2020 have been used. Each record in the datasets denotes a single locality of an infection case in a certain specific week. For the outbreak year 2020, we collect the dengue locality data from 22 Feb to 3 July, which covers the whole period of the nationwide lockdown (known as the circuit breaker) for controlling the pandemics of COVID-19 as well as a period of time before and after it respectively.
We define a subzone as a hotspot if it contains no less than a certain number (denoted as $c$) of localities in a given week. Let $l_i^t$ denote the number of the infection localities in subzone $i$ for a given week $t$. The corresponding hotspot state is defined as

\begin{equation}\label{eq:hotspot}
    \mathbf{y}_i^t=\left\{
    \begin{array}{cc}
         1, &  l_i^t\geq c\\
         0, & otherwise
    \end{array}
    \right.
\end{equation}

\subsection{Hotspot Dynamics Modeling}

 In the real situation, the regions that do not meet the hotspot threshold could also exert a potential impact on future hotspots. To take them into consideration, another state variable $\mathbf{\hat{y}}_i^t$ is defined in the modeling of the hotspot dynamics where $\mathbf{\hat{y}}_i^t=1$ if $l_i^t\geq1$; otherwise $\mathbf{\hat{y}}_i^t=0$. 
The process of dengue hotspot dynamics modeling is schematically shown in Fig. 1. Urban regions may become hotspots along the time axis, and such temporal dynamics can be further viewed from the spatiotemporal dimension, where the virus spreads across different regions.

It is known that dengue’s main vector (Aedes aegypti), are highly anthropophilic, aggregating in and around human premises and dispersing relatively short distances (a few hundred meters) \cite{b21}. Thus, it is generally assumed that human movements contribute to the transmission of dengue on spatial scales that exceed the limits of mosquito dispersal \cite{b7}, carrying dengue virus to previously dengue-free areas and infecting local mosquitoes \cite{b26}. Infected mosquitoes’ biting further transmits the dengue virus to local residents  \cite{b27}. Driven by human movements, dengue virus may quickly disperse across cities or urban areas \cite{b28,b29,b30,b30b}. The subzones used in this study averagely cover 1.35 $km^2$, which in most cases exceeds the mosquito fly range. As the inter-subzone transmission is mainly due to human travel, it is reasonable to assume that the state $\mathbf{y}_i^t$ is associated with the epidemic dynamics of all subzones (including the target subzone $i$ itself) in previous weeks. The ultimate objective of the algorithm design is to utilize such latent association involved in the hotspot dynamics to forecast dengue hotspots in the next week, namely, to forecast the hotspot state  $\mathbf{y}_i^{t+1},i\in{1,2,…,N}$, for all subzones.

Given the current week t, we utilize a matrix, $\mathbf{P}^t$, to model the hotspot dynamics, with each matrix element indicating the underlying association of dengue spreading between two subzones, \textit{i.e.}, the dash lines in Fig. \ref{fig:overview}. Note that, as the transmission routes are invisible, the entities in the matrix, $\mathbf{P}_{ij}^t$, which denotes the weight of the spreading link from subzone $j$ to subzone $i$, is initially unknown. 

In addition, the epidemic dynamics in subzone $j$ during two different weeks $t-h$ and $t-h'$ may exert distinct influences on the current hotspot state in subzone $i$, \textit{i.e.}, $\mathbf{y}_i^t$. Hereby, a temporal parameter, $w_h^t$, is introduced. The temporally weighted combination of $\mathbf{\hat{y}}^(t-h)$ in the past $H$ weeks is then represented as $\sum_{h=1}^{H}w_h^t\mathbf{\hat{y}}^{t-h}$.

By further incorporating $\mathbf{P}^t$ into the temporal combination, we model the hotspot estimations as $\sum_{h=1}^{H}w_h^t\mathbf{P}^t\mathbf{\hat{y}}^{t-h}$ for the current week $t$, where $\mathbf{P}^t\in\mathbb{R}^{N\times N}$ is the spatiotemporal parameter and $w_h^t\in[0,1]$ is the temporal parameter. 

At this point, we have the hotspot estimations of all subzones for the current week $t$, represented as $\sum_{h=1}^{H}w_h^t\mathbf{P}^t\mathbf{\hat{y}}^{t-h}$. As well, we know the ground truth of the current hotspot states of all subzones, \textit{i.e.}, $\mathbf{y}^t$. What we intend to do is to learn the two parameters $\mathbf{P}^t$ and $w_h^t$ in the spatiotemporal pattern learning process based on the hotspot observations in previous weeks $t-H,...,t-1$ as well as in the current week $t$. Then we can utilize the learnt parameters to forecast the future hotspots for the next week $t+1$, \textit{i.e.}, to determine the values (0 or 1) for $\mathbf{y}_i^{t+1},i\in\{1,2,…,N\}$. 

The framework of the proposed method can be briefly summarized as follows. The national dengue dynamics from the temporal view is decomposed into the regional hotspot dynamics along the timeline. The current hotspot state of a region is correlated to previous hotspots states of all subzones. The model consists of the spatiotemporal pattern learning and the future hotspot forecasting, respectively. In the learning process, spatiotemporal parameter $\mathbf{P}^t$ are learnt by a gradient descent algorithm using the hotspot data in weeks $(t-h,…,t)$, where $t$ denotes the current week. Detailed learning method for $\mathbf{P}^t$ shall be described in the next section. With the learned parameters and using the hotspot data in the recent weeks $(t-h+1,…,t)$, the one week ahead $(t+1)$ hotspot states of all regions can be predicted. 

Specifically, for the hotspot forecasting, we utilize the learnt $\mathbf{P}_{ij}^t$ and $w_h^t$, together with $\mathbf{\hat{y}}^t,…, \mathbf{\hat{y}}^{t-H}$, to forecast the one-week-ahead hotspot state, $\mathbf{y}^{t+1}$, which is calculated as
\begin{equation}
    \mathbf{y}^{t+1}=\mathcal{J}\left(\mathcal{A}\left(\sum_{h=1}^H w_h^t\mathbf{P}^t\mathbf{\hat{y}}^{t+1-h}\right)\right)
\end{equation}

where $\mathcal{A}=tanh$ is an activation function, 
$\mathcal{J}(x)$ is an indicator function with $\mathcal{J}(\mathbf{x}_i)=1$ if $\mathbf{x}_i >\mu_x+\sigma_x$, and 0 otherwise.

\subsection{Spreading Pattern Learning}
The hotspot estimation for the current week $t$, based on hotspot observations from the preceding $H$ weeks, is modeled as

\begin{equation}
    \sum_{h=1}^{H}w_h^t\mathbf{P}^t\mathbf{\hat{y}}^{t-h}
\end{equation}

where $w_h^t$ denotes the temporal weighting of week $t-h$ and $\mathbf{P}^t$ is the latent spatial spreading pattern to be learned in the recent $H$ weeks, both of which are learnable model parameters. For a given week $t$, both parameters could be learned by minimizing errors between the hotspot estimations and the observations. For the temporal weight $w_h^t$, we search the best value by enumerating from zero to one with a small increment for each iteration, such as 0.01. The latent spatial spreading matrix $\mathbf{P}^t$ can be learned by gradient descent given the objective loss function that is the gap between the hotspot estimations and the observations.

Given a group value of ${w_1^t,w_2^t,…,w_H^t}$, let $\mathcal{L}^t$ be the global loss between the hotspot estimations, $\sum_{h=1}^{H}w_h^t\mathbf{P}^t\mathbf{\hat{y}}^{t-h}$, and observations, $\mathbf{y}^t$. It is the objective function to be minimized and is calculated as the following equation. 
\begin{equation}
\begin{split}
   \mathcal{L}^t=\sum_{i=1}^N\Vert \sum_{h=1}^H \left(w_h^t\mathbf{P}_{i\cdot}^t\mathbf{\hat{y}}^{t-h}-\mathbf{y}_i^t\right) \Vert _2^2 \\
    + \lambda_1\Vert\mathbf{P}^t\Vert_2^2+\lambda_2\Vert\mathbf{P}^t\Vert_1 
\end{split}
\end{equation}

where $\lambda_1\Vert\mathbf{P}^t\Vert_2^2$ is the $L_2$ norm regularization term that makes the model acquire small values in $\mathbf{P}^t$, and $\lambda_2\Vert\mathbf{P}^t\Vert_1$ is the $L_1$ normal regularization term that makes the model acquire a sparse matrix $\mathbf{P}^t$, meaning a sparse matrix $\mathbf{P}^t$, \textit{i.e.}, most of matrix elements approximate to zero. 

For learning $\mathbf{P}^t$, the partial gradient $\partial{\mathcal{L}^t}/\partial{\mathbf{P}^t}$ is calculated by rows. Specifically, for the $i^{th}$ row of $\mathbf{P}^t$, \textit{i.e.}, $\mathbf{P}_{i\cdot}^t$, the local loss is represented as $\mathcal{L}_i^t$ and then the partial derivative $\partial{\mathcal{L_i}^t}/\partial{\mathbf{P}^t_{i\cdot}}$ is calculated as

\begin{equation}
\begin{split}
    \frac{\partial{\mathcal{L}_i^t}}{\partial{\mathbf{P}_{i\cdot}^t}}=2\sum_{h=1}^H\left(w_h^t\mathbf{P}_{i\cdot}^t\mathbf{\hat{y}}^{t-h}-\mathbf{y}_i^t\right)\times \sum_{h=1}^H\left(w_h^t\mathbf{\hat{y}}^{t-h}-\mathbf{y}_i^t\right)^T 
\\
    +\lambda_1\mathbf{P}_{i\cdot}^t + \lambda_2sign(\mathbf{P}_{i\cdot}^t)
\end{split}
\end{equation}

The objective function could be optimized by using the gradient descent algorithm. Specifically, $\mathbf{P}^t$ is initially randomized. Subsequently, in each iteration, a negative gradient, $-\partial{\mathcal{L}_i^t}/\partial{\mathbf{P}^t_{i\cdot}}$, is added to the $i^{th}$ row of $\mathbf{P}^t$ for updating. By repeating this process, the matrix $\mathbf{P}^t$ could be learned.

Dengue hotspot forecasting for week t+1 is to determine the values (0 or 1) for $\mathbf{y}_i^{t+1},i\in\{1,2,…,N\}$ with the given hotspot observations. The forecasting algorithm is shown in the \textit{Appendix B.}
The hyperparameters of the model, $H,\lambda_1,\lambda_2$, are set by cross variation to select the values for best forecasting. In our case studies of Singapore dengue spreading modeling, the values are selected as $H=4,\lambda_1=0.01,\lambda_2=0.1$.

\subsection{Interpretability of Spatiotemporal Parameter $\mathbf{P}^t$}

Note that the proposed model incorporates the spatial transmission links among urban regions as a model parameter for each week, \textit{i.e.}, $\mathbf{P}^t$. By aggregating these learnt matrices for each year, we can have the yearly transmission links between urban regions. Specifically, we obtain the yearly spatial spreading network from the learnt matrix as

\begin{equation}
    \mathcal{G}_{\mathcal{L}}(i,j)=\sum_{t=1}^{T}\frac{\mathbf{P}^t(i,j)}{
    \max _{1\leq i,j\leq N}\mathbf{P}^t(i,j)}
\end{equation}

Here, $T$ is the total number of weeks in a year.  The matrix value is first divided by $\max _{1\leq i,j\leq N}\mathbf{P}^t(i,j)$ before further aggregation, which is to homogenize the matrices, \textit{i.e.}, to ensure that the maximum value of each element is 1. We take $\mathcal{G}_{\mathcal{L}}(i,j)$ as the yearly spatial spreading network of dengue across subzones, which reveals the spreading ratio to subzone $i$ from subzone $j$ on a relatively long-term basis.

The study in \cite{b31} demonstrates that human commuting flows contribute to the dengue spreading across regions, which exceeds the flying range of the mosquitoes. We obtain the human mobility flows among subzones from the mobile phone data in 2011 used in \cite{b31}. Note that the aggregated network, $\mathcal{G}_\mathcal{L}$, is learnt by the model based on only the dengue hotspot observations with no information of the mobility networks. In this study, we shall compare the learnt transmission network, $\mathcal{G}_\mathcal{L}$, with the human commuting flows in Singapore to reveal the correlation between them.

We first describe the mobility network with the dataset used in \cite{b31}. The study in \cite{b31} extracted the anonymized mobile users’ home and work localities in 320m$\times$320m grids and calculated the mobility flow between these grids. The home and work locations of a total of 2,307,230 anonymized mobile users are assigned to the corresponding grids. 

Hence each record is a tuple $(g_{i'}^m,g_{j'}^m)$, where $g_{i'}^m$ is the home location grid of a user m, and $g_{j'}^m$ is the user’s work location grid. 
In this study, we map each grid to the corresponding subzone. Thus, we have a tuple $(z_i^m,z_j^m)$ for each user $m$.  By aggregating the commuting flows between each pair of subzones, we have the mobility networks denoted as $\mathbf{\mathcal{G}}_\mathcal{M}$, where $\mathbf{\mathcal{G}}_\mathcal{M}(i,j)$ is the mobility flow between subzones $i$ and $j$. 
Note that we let $(z_i^m,z_j^m)$ contribute a mobility flow to both $(i,j)$ and $(j,i)$. Hence $\mathbf{\mathcal{G}}_\mathcal{M}$ is a symmetric graph. In our study, $\mathbf{\mathcal{G}}_\mathcal{M}$ is generated based on the commuting flows in Singapore for year 2011. We shall compare $\mathbf{\mathcal{G}}_\mathcal{L}$ and $\mathbf{\mathcal{G}}_\mathcal{M}$ to test on the consistency between them. Note that, the mobility network $\mathbf{\mathcal{G}}_\mathcal{M}$ is only available for year 2011. We assume that the citywide mobility pattern in 2011 remained largely the same for comparing $\mathbf{\mathcal{G}}_\mathcal{M}$ with $\mathbf{\mathcal{G}}_\mathcal{L}$ in 2013.

\section{Case Study and Analysis}
\subsection{Study Area and Spatial Unit}
\begin{figure}[ht]
    \centering
    \includegraphics[width=0.48\textwidth]{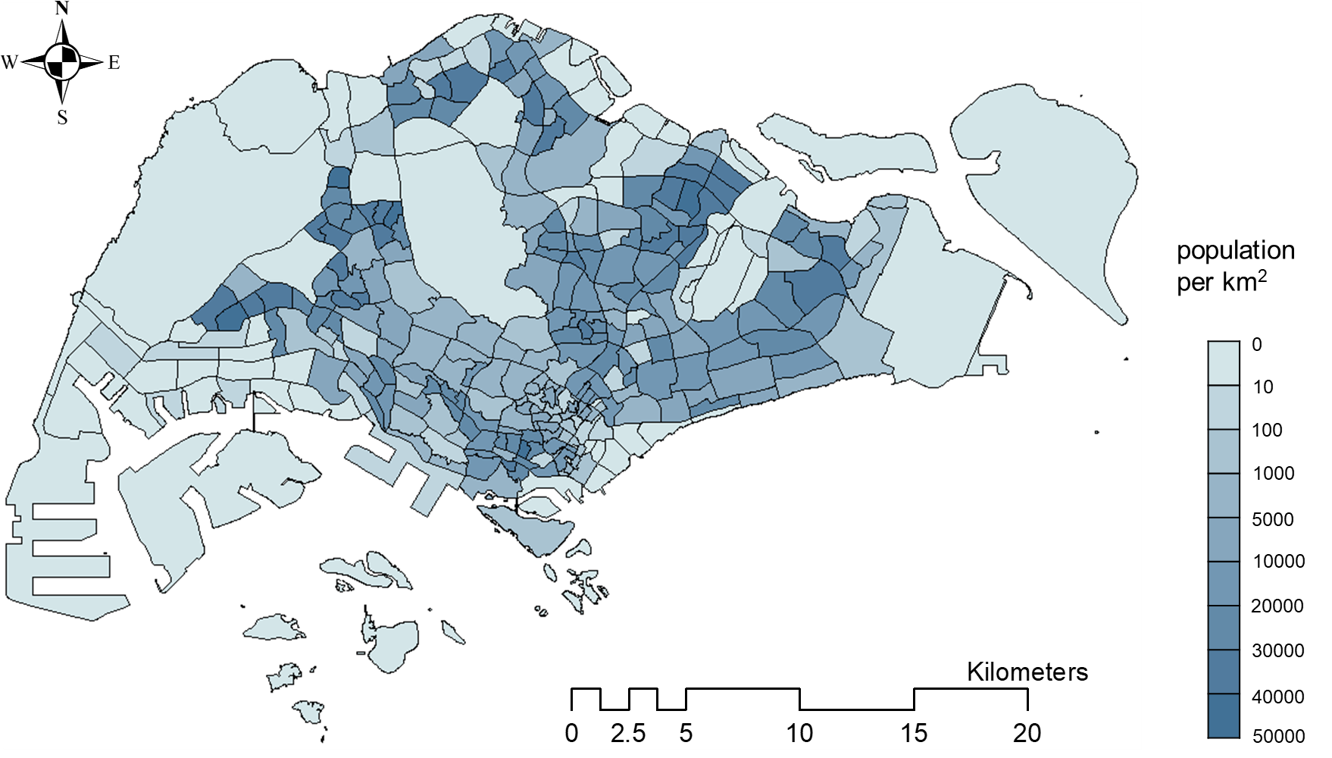}
    \caption{Visualization of the Urban Subzones in Singapore with Corresponding Population Density.}
    \label{fig:subzone}
\end{figure}

The city-state of Singapore lies about one degree of latitude north of the equator (1.290270° N; 103.851959° E), covering a total of 781.9 km2 and accommodating 5.7 mil-lion residents. The urban redevelopment authority (URA) of Singapore divides the nation into 323 subzones \cite{b32}, each of which is typically centered around a focal point such as a neighborhood center. We utilize the sub-zones as the basic spatial units for dengue hotspot forecasting as the research in \cite{b33}. Except for those subzones whose population densities are lower than 10 per $km^{2}$, the mean coverage of the subzones is only about 1.35 $km^2$, which provides a relatively fine-grained spatial unit. On a subzone that is predicted to be likely a hotspot in the next week, NEA may invest more extensive vector control efforts. We visualize the urban subzones in Singapore in Fig. \ref{fig:subzone}, where the sub-zones are color-coded by the population density. 
In this work, we set the threshold $c=3$ in Eq. \ref{eq:hotspot}. As each subzone only covers an average of 1.35 $km^{2}$, having three or more localities with reported dengue cases implies that the subzone is of high transmission risk and possibly with intracommunity transmissions between close-by buildings. Applying NEA’s mosquito control in such subzones may be effective.

\subsection{Hotspot Modeling Evaluation}
Prior to analyzing the mined dengue transmission network across urban regions, it is essential to confirm the reliability of the dengue hotspot dynamics model, since explainable parameters can only be derived from a well-validated framework.
The forecasting results are shown in Table 1. As can be seen, results on the dengue case dataset of Singapore for years 2013-2018 and 2020 show that the mean Accuracy values vary from 0.9219 to 0.9988 in different years, the mean Precision values vary from 0.7744 to 0.9611, the mean Recall values vary from 0.6557 to 0.8833, and the mean F-Score values vary from 0.7127 to 0.9010. With such satisfactory performance, the proposed hotspot dynamics modeling method may provide us valuable knowledge of the geographical dengue risk distribution at a citywide scale, helping prevent further disease transmission.
Taking only the dengue hotspot records as model inputs makes the model easy to be used for hotspot prediction. The acceptable forecasting performance for the special year of 2020 to a certain extent demonstrates the robustness of the proposed model. Modeling code can be found on GitHub at  \url{https://github.com/lphuang2022/dengue-hotspot-forecasting-and-spreading-network-recognition}.

\begin{table}[h]
    \centering
    \setlength{\tabcolsep}{6pt} 
    \renewcommand{\arraystretch}{1.2}
    
    \begin{tabular}{c|cccc}
    \hline
        
        Year &  Accuracy ($\mu, \sigma$)  & Precision & Recall  & F-1 \\
         \hline
         \hline
   \rowcolor{maroon!5} 2013 &  0.9771, 0.0134 & 0.7956 & 0.6557 & 0.7127\\
        \hline
                        2014 &  0.9376 , 0.0194 & 0.7848  & 0.7207  & 0.7398\\
        \hline
    \rowcolor{maroon!5} 2015 &  0.9637 , 0.0162 & 0.7744  & 0.7374  & 0.7526 \\
        \hline
                        2016 &  0.9699 , 0.0135 & 0.7800  & 0.7074  & 0.7320 \\
        \hline
    \rowcolor{maroon!5} 2017 &  0.9988 , 0.0016 & 0.9611  & 0.8833  & 0.9010\\
        \hline
                        2018 &  0.9969 , 0.028 & 0.9488  & 0.8260  & 0.8614\\
        \hline
    \rowcolor{maroon!5} 2020 &  0.9219 , 0.0145 & 0.8626  & 0.8007  & 0.8294\\
        \hline

    \end{tabular}
    \caption{Dengue Hotspot Forecasting Performance for Seven Years in Singapore}
    \label{tab:predict}
\end{table}

\begin{figure}[h]
    \centering
    \includegraphics[width=.38\textwidth]{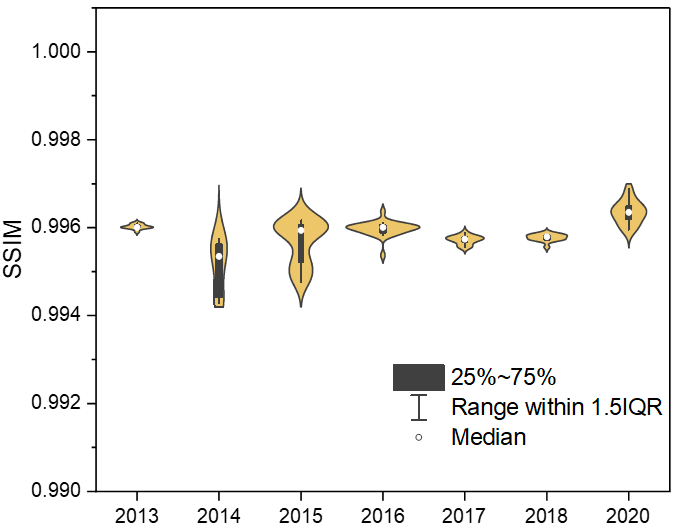}
    \caption{SSIM of the Learnt Matrix for Each Year}
    \label{fig:ssim}
\end{figure}

\begin{figure*}[bt]
    \centering
    \includegraphics[width=1\textwidth]{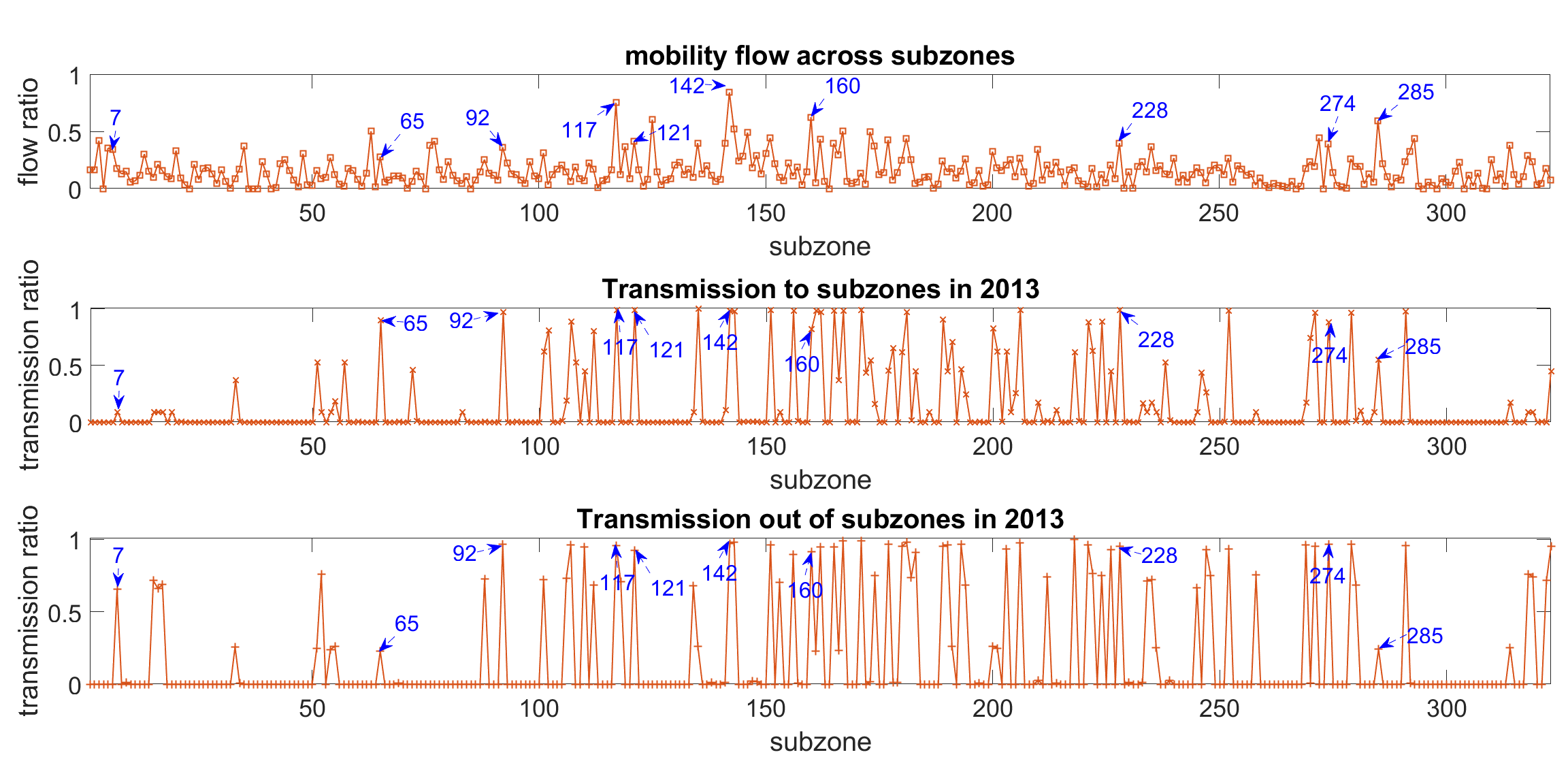}
    \caption{Mobility Flow Ratio (top), Transmission Ratio to Subzones (middle), and Transmission Ratio out of Subzones (bottom)}
    \label{fig:mob_tranR}
\end{figure*}

\begin{figure}[b]
    \centering
    \includegraphics[width=0.45\textwidth]{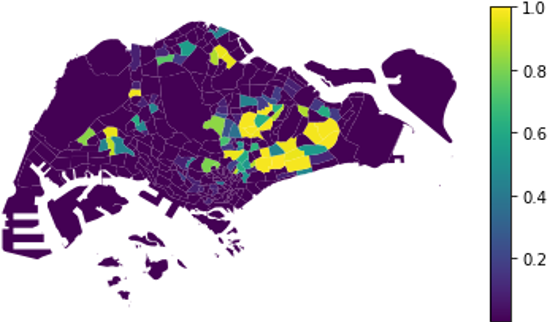}
    \caption{Geographical Distribution of Mobility Ratios}
    \label{fig:geo_mob}
\end{figure}

\begin{figure}[b]
    \centering
    \includegraphics[width=0.45\textwidth]{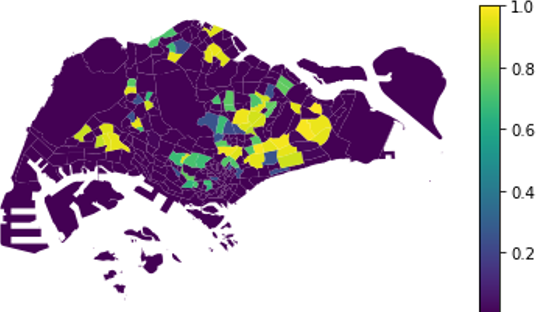}
    \caption{Geographical Distribution of Transmission-in}
    \label{fig:geo_in}
\end{figure}

\subsection{Stability of the Learned $\mathbf{P}^t$}
The hotspot prediction model incorporates the spatial correlation between subzones as a dynamic matrix $\mathbf{P}^t$. Note that $\mathbf{P}^t$ and $\mathbf{P}^{t+1}$ are independently learned by the gradient descent algorithm. Relative stability analysis for each pair of $\mathbf{P}^t$ and $\mathbf{P}^{t+1}$ can help us understand the temporal dynamics of matrix  $\mathbf{P}^t$. 

We utilize the Structure SIMilarity (SSIM) to analyze the similarity between $\mathbf{P}^t$ and $\mathbf{P}^{t+1}$, which is originally proposed for measuring the similarity between two images \cite{b31}. The range of the SSIM index value is between 0 and 1: when two metrics are nearly identical, their SSIM is close to 1. Before calculating SSIM, we first normalize $\mathbf{P}^t$ by row to homogenizing $\mathbf{P}^t$ in each week. Then, a higher value of SSIM between $\mathbf{P}^t$ and $\mathbf{P}^{t+1}$ denotes a stronger stability of $\mathbf{P}^t$. In the experiments, we calculate the SSIM between  $\mathbf{P}^t$ and $\mathbf{P}^{t+1}$  for each pair of adjacent epidemiological weeks of a given year. 
The SSIM values for each year are shown in Fig. \ref{fig:ssim}. It could be observed that the values of SSIM approximate to 0.96 or larger values, which means that the spatial matrix $\mathbf{P}^t$ is quite stable between two continuous weeks. 

\begin{figure*}[t]
    \centering  
    \includegraphics[width=1\textwidth]{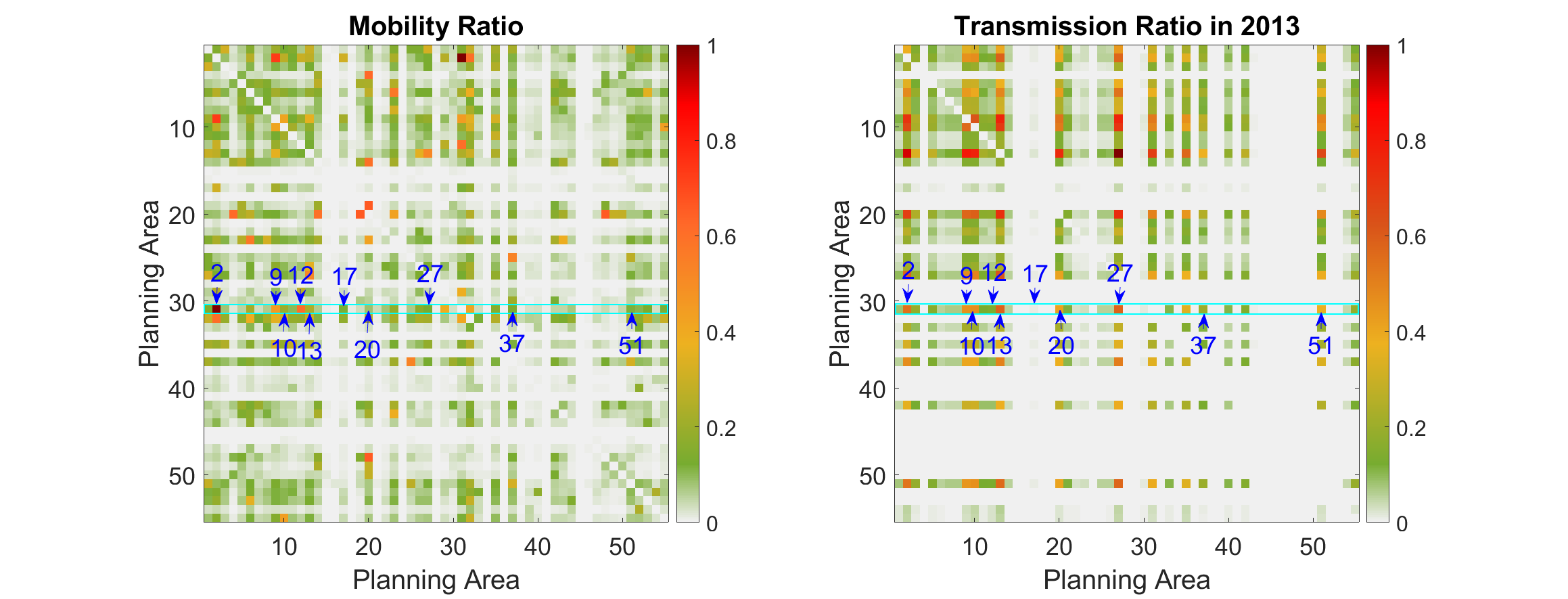}
    \caption{Mobility Ratios and Transmission Ratios to Planning Area 31.}
    \label{fig:viz_pa1}
\end{figure*}

\subsection{Interpreting the Spatial Spreading Network}

In this section, we compare the learned spatial spreading network $\mathcal{G}_\mathcal{L}$ of 2013 with the commuting mobility network, $\mathcal{G}_\mathcal{M}$. Note that the high diagonal values reflect a large number of local dengue transmissions which however may be caused by short-distance travels of the infected mosquitoes. In this section, we focus on inter-regional (non-diagonals) parts in both the mobility network, $\mathcal{G}_\mathcal{M}$, and the spatial spreading network $\mathcal{G}_\mathcal{L}$.

To facilitate analyzing the learned spreading network and comparing it with the mobility network, for each subzone $k$, we further define following three metrics, transmission-in $\mathcal{G}_{\mathcal{L},in}$, transmission-out $\mathcal{G}_{\mathcal{L},out}$ and mobility ratio $\mathcal{G}_{\mathcal{MD}}$, which are respectively calculated as following equations. $\mathcal{D}(k)$ is the population in subzone $k$.

\begin{equation}
    \mathcal{G}_{\mathcal{L},in}=\frac{\mathcal{G}_{in}}{max(\mathcal{G}_{in})}, 
    \mathcal{G}_{in}(k)=\sum_{j=1}^N\mathcal{G}_{\mathcal{M}}(k,j)
\end{equation}

\begin{equation}
    \mathcal{G}_{\mathcal{L},out}=\frac{\mathcal{G}_{out}}{max(\mathcal{G}_{out})}, 
    \mathcal{G}_{out}(k)=\sum_{j=1}^N\mathcal{G}_{\mathcal{M}}(j,k)
\end{equation}

\begin{equation}
    \mathcal{G}_{\mathcal{MD}}=\frac{\mathcal{G}_{\mathbf{D}}}{max(\mathcal{G}_{\mathcal{D}})}, 
    \mathcal{G}_{\mathbf{D}}=\sum_{j=1}^N\mathcal{G}_{\mathcal{M}}(k,j)\mathcal{D}(k)
\end{equation}

As shown in Fig. \ref{fig:mob_tranR}, we pick ten subzones, which have high $\mathcal{G}_{MD}$ values as well as high $\mathcal{G}_{\mathcal{L}, in}$ and $\mathcal{G}_{\mathcal{L}, out}$, for comparisons. The geographical distribution of the mobility density ratios $\mathcal{G}_{MD}$ and the transmission-in ratios, $\mathcal{G}_{\mathcal{L}, in}$ are shown in Fig. \ref{fig:geo_mob} and Fig. \ref{fig:geo_in}, which demonstrates the distribution similarities of the high values, especially in the east areas.  Note that the local mobility flows within a subzone are excluded, \textit{i.e.}, $\mathcal{G}_\mathcal{M}(i,i)$ are set to be zero. This is because these values are high and make other values difficult to observe in the same figure. For the learned spreading network, it can be observed that the diagonals, $\mathcal{G}_\mathcal{L}(i,i)$, are also of high values. These high values reflect a large number of local dengue transmissions, which, however, may be caused by short-distance travels of the infected mosquitoes.

For further visualization, the subzone-based graph $\mathcal{G}_\mathcal{M}$ and $\mathcal{G}_\mathcal{L}$ are aggregated into planning areas, which are also defined by URA and are utilized in \cite{b31}. Specifically, a total of 55 planning areas cover the entire region of Singapore. The mobility density ratios, and the transmission-in ratios  for year 2013 are shown in Fig. \ref{fig:viz_pa1}. We identify ten planning areas, namely 2, 9, 10, 12, 13, 17, 20, 27, 30, 37, 51, which are of high mobility flows to and from planning area 31.

From the figure on the right of Fig. \ref{fig:viz_pa1}, we find that these planning areas also have high transmission risks to and from the planning area 31. Such consistency to a certain extent indicates that the learned spreading network has significant similarities with the mobility flows, although the learning process does not make use of any mobility data. 

\subsection{Temporal Weights Analysis}
\begin{figure}[b]
    \centering
    \includegraphics[width=.45\textwidth]{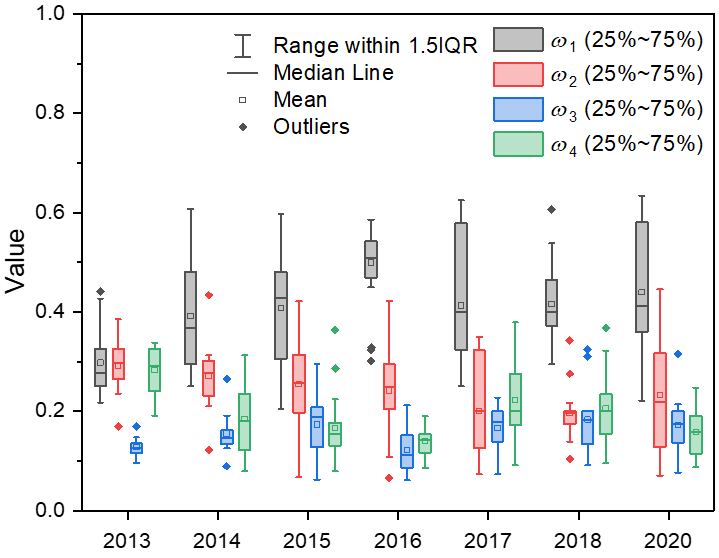}
    \caption{Temporal Weights Distribution for Each Year}
    \label{fig:tem_weights}
\end{figure}

The temporal weights $w_1^t, w_2^t, w_3^t, w_4^t$ are also analyzed to better understand the evolution of dengue risk over time. As shown in Fig. \ref{fig:tem_weights}, the magnitudes of these weights vary across years, highlighting temporal heterogeneity in how past hotspots influence current conditions. Nevertheless, a consistent trend emerges: $w_1^t$ and $w_2^t$ are generally larger than $w_3^t$ and $w_4^t$. This implies that dengue hotspots in the most recent one to two weeks have the strongest influence on the present hotspot state of a subzone. Hotspots from three to four weeks ago still contribute, but their impact is weaker, though not negligible. These results suggest that while dengue risk decays with temporal distance, early signals remain relevant within a month-long horizon—a property that can inform both predictive modeling and timely intervention strategies.

\section{Discusstion}
\subsection{Dengue Hotspot Dynamics}
In Singapore, NEA dynamically updates the information on dengue hotspots, which helps guide the vector control intervention. However, being able to forecast active transmission localities, rather than only knowing where they are currently, may allow mitigating mosquito density in advance, which would help prevent further virus transmission.

In this study, we propose a general model, at a city-wide level, that is capable of forecasting dengue hotspots. The spatial unit used in this study is the subzone, which is quite small with an average size of 1.35 $km^{2}$ and could help anchor the targeting areas for proactive vector control. Since whether a subzone turns to be or remains as a hotspot in the current week is collaboratively determined by epidemic dynamics of all subzones in the past a few weeks, the developed model incorporates such spatiotemporal impacts. By minimizing the errors between estimations and observations of hotspots in the current week with a gradient descent algorithm, the dynamic impacts are learned, with which the one-week-ahead hotspots are forecast. 

The proposed model takes only the dengue surveillance data in the previous four weeks as inputs, which is available in many other dengue endemic cities or countries. The proposed model hence may be applied to other urban areas with appropriate adjustments, \textit{e.g.}, to the spatial units and the threshold for defining a hotspot, etc. 

\subsection{Spreading Network Recognition}
The warm and rainy climate in the urban environments in tropical or sub-tropical nations provide perfect breeding beds for dengue vectors, and the intensive daily commuting flows in urban environments expedite the citywide spread of dengue virus. Though many studies have been carried out on the spatial spread of vector-borne disease at different scales, modeling the hotspot dynamics across regions in an endemic city remains a challenge due to the invisible infection pathways. Hence, a merit of the proposed model is that it captures both the spatial and temporal dynamics of dengue hotspots on a citywide scale. The unobservable transmission routes are set as the spatial correlation, $\mathbf{P}^t$, which is found to remain rather stable between continuous weeks. The comparisons between the spreading network with the commuting flows in Singapore may help us better understand the geographically structured dengue transmission in this city.

It is worth noting that in the year 2020, with COVID-19 outbreak, Singapore has been experiencing an unprecedented dengue outbreak at the same time. What makes the case of this year’s dengue spreading even more special is that the spike in early May coincided with Singapore’s COVID-19 lockdown period known as Circuit Breaker (CB), from 7 April to 1 June. Revealing how the CB measurements impact the dengue spreading is pivotal, especially for identifying the correlation between the commuting flows and dengue spreading across urban areas.

\section{Conclusion}
In conclusion, this study proposes a dynamics modeling of dengue hotspots to recognize the dengue spreading among urban regions. The case study in 
Singapore shows that the dengue hotspot dynamics can be captured by the proposed model at an accuracy. Moreover, the spatial spreading of dengue hotspots in a highly urbanized environment can be recognized by the proposed method, which has been verified via comparing the recognized network with human mobility among uban regions. The model requires only the geolocated dengue case data within the current and previous four weeks, which are available or can be made available in many dengue endemic cities or nations. Such kinds of findings indicate that the proposed method are suitable for routinely guiding vector control efforts. The proposed method, with appropriate adjustments, may help enhance dengue control in other urban areas.




\begin{acks}
Flora Salim would like to acknowledge the support of the ARC Center of Excellence for Automated Decision Making and Society (CE200100005) and the CSIRO-NSF grant titled "Collaborative Research: NSF-CSIRO: HCC: Small: Understanding Bias in AI Models for the Prediction of Infectious Disease Spread”. This work is also supported in part by the Key R$\&$D Project of Jilin Province (No. 20240304200 SF).
\end{acks}

\bibliographystyle{ACM-Reference-Format}
\balance
\bibliography{web4good}










\end{document}